# Achieving Domain Generalization for Underwater Object Detection by Domain Mixup and Contrastive Learning

Yang Chen[a,b], Pinhao Song[a], Hong Liu[a,*], Linhui Dai[a], Xiaochuan Zhang[b], Runwei Ding[c] and Shengquan Li[c]

[a]Key Laboratory of Machine Perception, Shenzhen Graduate School, Peking University, Shenzhen, 518055, Guangdong, China
[b]School of Artificial Intelligence, Chongqing University of Technology, Chongqing, 401135, China
[c]Peng Cheng Laboratory, Shenzhen, 518038, China

ABSTRACT

The performance of existing underwater object detection methods severely degrades when they face the domain shift caused by complicated underwater environments. Due to the limited domain diversity in collected data, deep detectors easily memorize a few seen domains, which leads to low generalization ability. There are two common ideas to improve the domain generalization performance. First, it can be inferred that the detector trained on as many domains as possible is domain-invariant. Second, their hidden features should be equivalent because the images with the same semantic content are in different domains. This paper further excavates these two ideas and proposes a domain generalization framework that learns how to generalize across domains from Domain Mixup and Contrastive Learning (DMCL). First, based on the formation of underwater images, an image in one kind of underwater environment is the linear transformation of another underwater environment. Therefore, a style transfer model, which outputs a linear transformation matrix instead of the whole image, is proposed to transform images from one source domain to another, enriching the domain diversity of the training data. Second, the Mixup operation interpolates different domains on the feature level, sampling new domains on the domain manifold. Third, a contrastive loss is selectively applied to features from different domains to force the model to learn domain-invariant features but retain the discriminative capacity. With our method, detectors will be robust to domain shift. Also, a domain generalization benchmark S-UODAC2020 for detection is set up to measure the performance of our method. Comprehensive experiments on S-UODAC2020 and two object recognition benchmarks (PACS and VLCS) demonstrate that the proposed method is able to learn domain-invariant representations and outperforms other domain generalization methods. The code is available in https://github.com/mousecpn/DMC-Domain-Generalization-for-Underwater-Object-Detection.git

## 1. Introduction

Object detection is a task that aims at identifying and localizing all objects of specific categories in an image. Object detection methods based on deep learning are also exploited in underwater robotic tasks, such as underwater robot picking, fish farming, and biodiversity monitoring.

Compared with conventional scenes, complicated underwater environments bring significant challenges to underwater object detection since domain shift often happens in underwater environments. For example, different times in one day cause different light conditions, and water qualities vary significantly in different areas, such as lakes and oceans. Moreover, due to the enormous difficulties of collecting and annotating underwater images, it is hard to integrate a large underwater dataset with rich domain diversity. In practical use, it is unrealistic for detectors trained on the dataset with limited domain diversity to adapt to the changing underwater environments, resulting from detectors suffering from domain shift. Therefore, it is indispensable to drastically improve the detectors' robustness to domain shift. The actual application in the wild can be seen as a domain generalization task in which the model is trained on a series of source domains and evaluated on an unseen but related domain. Most of the previous works focus on aligning source domains, such as Minimizing the Maximum Mean Discrepancy (MMD) [25] and using adversarial training [32]. Besides, current domain generalization works are devoted to recognition [23], but few works concern detection.

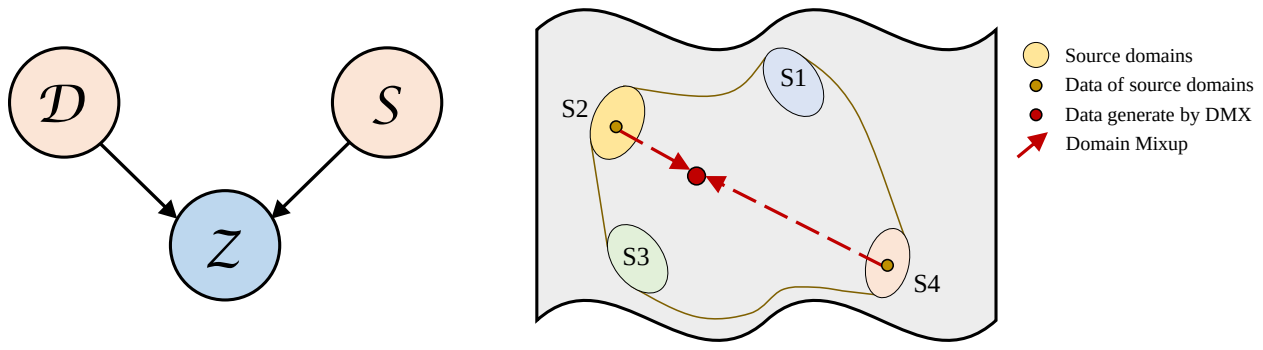

(a) Bayesian model (b) The latent domain manifold

**Figure 1:** The illustration of domain distribution. (a) $Z$ is the data distribution, $D$ is the domain distribution, and $S$ is the semantic contents distribution. (b) Each source domain occupies a very small area (the small ellipse in Fig. (b)), which can not represent the whole structure of the domain manifold. Domain Mixup can sample data inside the domain convex hull that is constructed by source domains.

*Corresponding author.
chenyang@pku.edu.cn (Y. Chen); Pinhaosong@pku.edu.cn (P. Song); hongliu@pku.edu.cn (H. Liu); dailinhui@pku.edu.cn (L. Dai); zxc@cqut.edu.cn (X. Zhang); dingrw@pcl.ac.cn (R. Ding); lishq@pcl.ac.cn (S. Li)
ORCID(s):

[1]This work is supported by the National Natural Science Foundation of China (NSFC, No.62073004) and Shenzhen Fundamental Research Program (JCYJ20200109140410340 and GXWD20201231165807007-20200807164903001).

Our method originates from two common assumptions: (1) increasing the sampling on the domain distribution helps to improve the robustness to domain shift. As Fig. 1 shows, the data distribution $\mathcal{Z}$ depends on two prior distributions: the domain distribution $\mathcal{D}$ and the semantic contents distribution $\mathcal{S}$. $\mathcal{D}$ constructs a latent domain manifold. A small number of domains in the dataset can not represent the structure of the domain manifold, leading to domain overfitting. If more domains can be sampled for training, the detector will succeed in eliminating the effect of domain shift. (2) An ideal feature extractor regards features from two different domains with the same semantic contents as equivalence, which is a straightforward idea. Further, simply penalizing the variance of the features will lead to the decline of the discriminative ability of the model. Therefore, the concepts of margin and selection are leveraged to allow room for the variance of the features to retain discriminative power.

To this end, Domain Mixup and Contrastive Learning (DMCL) is proposed to resolve the domain generalization problem in underwater object detection. First, the style transfer model named Conditional Bilateral Style Transfer (CBST) is proposed to transform an underwater image from one source domain to another. Second, Domain Mixup (DMX) on the feature level is proposed to interpolate two different domains to synthesize a new one. Both CBST and DMX increase the domain diversity of training data. Third, Spatial Selective Marginal Contrastive (SSMC) Loss is proposed to regularize the domain-specific features learned by the model. Finally, a new benchmark S-UODAC2020 is set up. S-UODAC2020 is a domain generalization dataset for the underwater object detection task, which indicates the robustness of detectors when facing the domain shift in the underwater environment. As shown in Fig. 2, S-UODAC2020 dataset contains 7 domains (*type1* to *type7*) with four classes: echinus, holothurian, starfish, and scallop. Each domain represents a kind of water quality. In S-UODAC2020, the detector is trained on *type1* to *type6* but evaluated on *type7*. As is shown, the gap between domains is large. Only the model truly learns the semantic contents can it perform well on the target domain. Several mainstream domain generalization works are transferred to our datasets for comparison.

In summary, the main contributions of this paper are summarized as follows:

- A style transfer model CBST is proposed to transform images from one source domain to another, enriching the domain diversity of the training data.

- A training paradigm named DMX is proposed to sample new domains on the domain manifold by interpolating paired images on the feature level.

- A regularization term SSMC loss is proposed to regularize the domain-specific features extracted by the backbone.

- A new dataset S-UODAC2020 is set up, which is designed for the domain generalization problem of underwater object detection. Comprehensive experiments not only show that our method achieves superior performance on domain generalization tasks, but also prove the effectiveness of the proposed components.

The following sections are structured as follows. Section 2 comprehensively analyzes the related works of the proposed methods. Section 3 describes the detailed architecture of the proposed method. Section 4 is the experiments and discussion, in which we conduct various experiments to demonstrate the effectiveness of the proposed method, analyze its characteristics, and prove the function of each component. Section 5 is the conclusion.

## 2. Related works

**Object Detection.** Existing object detection methods can be categorized into two groups: two-stage detectors and one-stage detectors. For two-stage detectors [42, 53, 2], the first stage is to use the Region Proposal Network (RPN) to propose candidate object bounding boxes. In the second stage, the detection head realizes the classification and bounding boxes regression of all objects based on the features extracted by CNN and RoI Pooling. One-stage detectors [37, 41] drop the RPN and the RoI Pooling, directly regressing the coordinates of bounding boxes and classes of objects for real-time processing.

The works mentioned above are all anchor-based methods. Anchor is a strong tool for object detection, for it degrades the detection task to the classification task, and the regression task based on the anchor is easier to train. Some researches focus on the design of anchor. MetaAnchor [59] uses anchor functions to generate anchors from the arbitrary customized prior boxes dynamically. GuidedAnchor [49] considers replacing the predefined set of scales and aspect ratios with an adjustable set that can be learned by the network. More radically, some works decide to abandon the use of anchor, such as FCOS [47], CornerNet [22] and CenterNet [11].

**Underwater Object Detection.** As an indispensable technology for AUVs to perform multiple tasks under the water, underwater object detection has attracted a large amount of attention from researchers all around the world. Existing underwater datasets are limited in quantity, and some works focus on data augmentation to increase the diversity of the dataset [19, 34, 35]. RoIMix [34] is a data augmentation method that applies Mixup on the RoI level to imitate occlusion conditions. Feature extraction is also important in underwater object detection. Several works find that increasing receptive field such as using dilated convolution helps improve the detection performance in underwater environments [13, 5]. For example, FERNet [13] proposed a receptive field augmentation module based on RFBNet [36]. Attention-based methods [33] and feature-pyramid-based methods [66] also improve the feature extraction capability in underwater environments. However, the method mentioned before does not pay attention to the domain shift in the underwater environment. DG-YOLO [35] is the first work focusing on the underwater domain shift problem and proposed the task of underwater object detection for domain generalization. Chen et al. [7] reveal that underwater image enhancement is beneficial to underwater detection in the wild by reducing the domain shift between training data and real-world scenes. In this work, unlike other works of underwater object detection, we study the domain generalization problem in the underwater environment as [35]. Through the ideas of domain sampling and semantic consistency, the proposed DMCL can achieve higher domain generalization performance than DG-YOLO. Besides, the experiments also demonstrate the limitation of common underwater object detectors in the domain generalization setting.

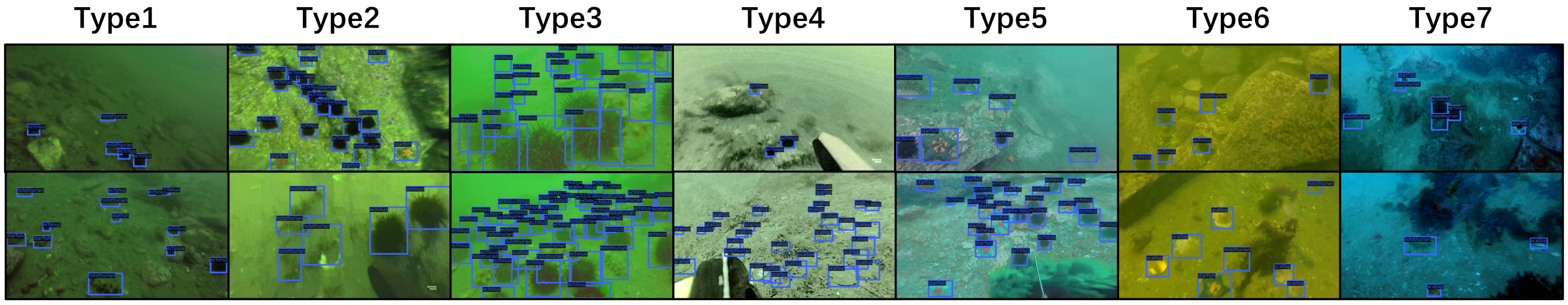

**Figure 2:** Some examples with ground truths in the proposed S-UODAC2020 dataset. S-UODAC2020 contains 7 domains (*type1* to *type7*).

**Domain Adaptation and Generalization.** In domain adaptation, a source domain dataset with ground truth and an unlabeled target domain dataset are available. It is hoped to fully use all these data during training and achieve good performance in the target domain. Generally, those two domains are related and share the same label space. Many works are proposed to solve domain adaptation in the recognition task. DANN [14] aligns features of source and target domains by adversarial training. Besides adversarial training, another solution is generating target domain data with source domain data by VAE [58] and GAN [16]. Further, some works focus on domain adaptation in the detection task [8, 43].

Domain generalization is a similar but more challenging task in which the model is trained on a series of source domains and evaluated on an unseen but related domain. There are five kinds of ideas to achieve domain generalization. First, some works focus on aligning features of the source domains [40, 32, 25]. For example, CCSA [40] proposes the semantic alignment loss to align features of the same class in different domains and the separation loss to separate features of different classes in the same domain. Second, data augmentation-based methods aim to synthesize data to enrich the domain diversity of training data [44, 67]. For example, CrossGrad [44] uses the concept of adversarial attack and defense, sampling domain-guided perturbation to augment the training data. Third, self-supervised training is used to increase the robustness to domain shift. JiGEN [3] leverages the self-supervised training paradigm Jigsaw Puzzle to improve the generalization capacity of the model. EISNet [50] not only leverages Jigsaw Puzzle, but also holds a memory bank for metric learning. Fourth, aggregation-based methods are proposed. D-SAM [12] trains domain-specific classifiers and makes decision fusion on all classifiers. Fifth, the meta-learning paradigm is also used for domain generalization, such as MetaReg [1] and MASF [10]. However, many domain generalization works stop at the recognition task, while domain generalization for object detection is understudied, which is what this paper focuses on.

**Real-time Style Transfer.** Gaty et al. [15] discover that the gram matrix of features extracted by ImageNet pre-trained VGG [46] is related to the style of an image, and a stylized image can be generated through iterative optimization. Early works on style transfer are optimization-based [15, 38], which is extremely time-consuming. [21] proposes a real-time style transfer model with Conditional Instance Normalization (CIN), which can generate multiple styles without training multiple times and achieve real-time processing. Besides, [20] proposes an arbitrary real-time style transfer with Adaptive Instance Normalization (AdaIN). Also, there is effort paid to photorealistic style transfer [29, 31, 61]. However, these photorealistic style transfer are also time-comsuming.

Recently, combining the real-time style transfer and photorealistic style transfer, Bilateral Style Transfer (BST) [56] proposes a photorealistic arbitrary real-time style transfer based on the design of HDRNet [26]. Unlike other end-to-end real-time style transfer methods, BST learns a color transform matrix from the style image and applies it to the content image instead of directly outputting the stylized image. There is a similarity between BST and the formation of underwater images: an image in an underwater environment is the linear transformation of an image in another underwater environment. Therefore, we decide to use BST as our basic style transfer model and modify it to propose CBST, which achieves faster inference speed and obtains clearer semantic contents.

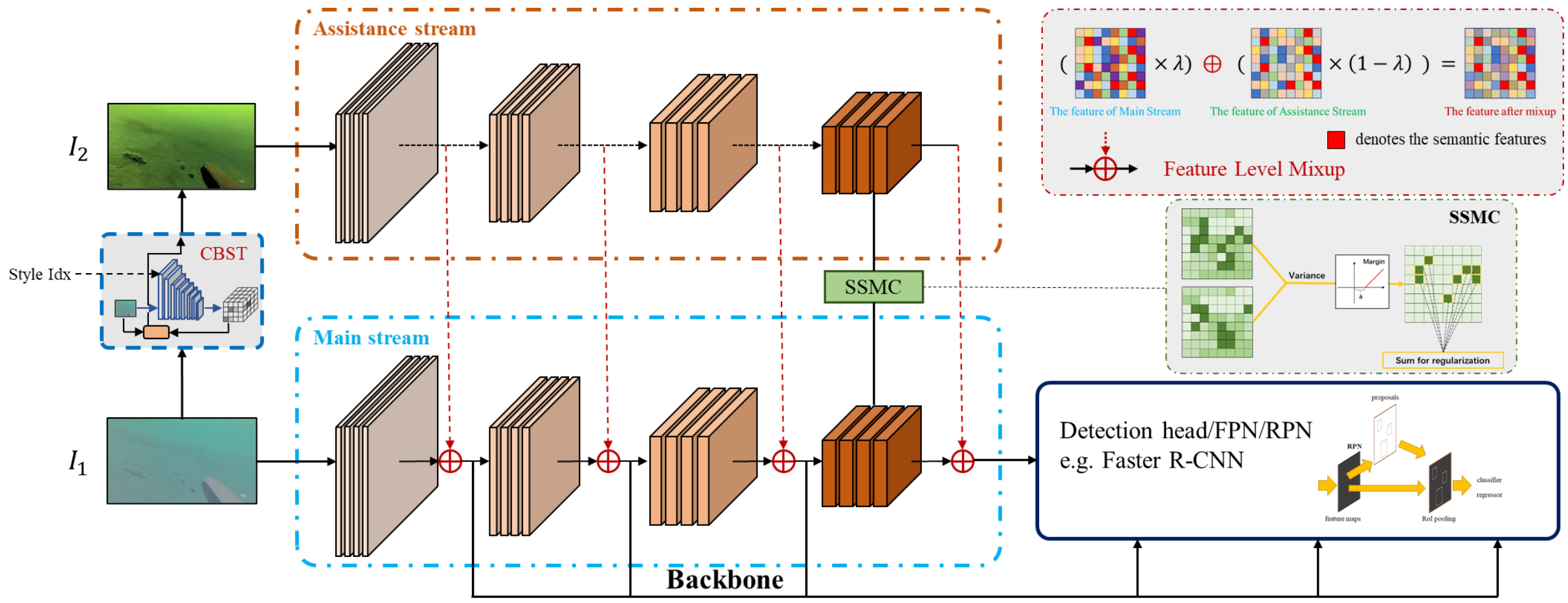

**Figure 3:** The overview of our proposed method. CBST converts an image to another source domain. As the paired images of different domains with the same semantic contents are obtained, they are fed into the backbone simultaneously. Domain Mixup is operated on certain layers of the backbone. The red pixels in the feature maps denote the semantic features, while others are the irrelevant domain features. SSMC loss is applied on the last convolutional layer to regularize domain-specific information of feature maps.

**Contrastive Learning.** Contrastive learning, whose goal is increasing the similarity between positive pairs and decreasing the similarity between negative pairs, has shown promising performance in self-supervised representation learning [18, 6, 55, 4, 17, 30]. In the past few years, there emerged numerous self-supervised representation learning works based on contrastive learning. For instance, MoCo [18] builds a momentum encoder and minimizes the distance between the data of two transformations against a large memory bank. SimCLR [6] uses various data augmentation and a large batch size to construct the positive and negative pairs. These methods have achieved advanced results and are easy to transfer to other areas, such as object detection [57], image segmentation [27, 51, 68], video reconstruction [28] and 3d object detection [60]. In this paper, we apply contrastive learning to the domain generalization problem in order to obtain semantic consistency between different domains.

## 3. The Proposed Method

The overview of our method is shown in Fig. 3. Our method contains three components: CBST, DMX, and SSMC loss. First, an image selected from one source domain is sent into CBST to be converted to another source domain. Second, since paired images are obtained, DMX is applied on the feature level during training to sample new domains inside the domain convex hull constructed by source domains. Third, SSMC loss is applied to the features of paired images to regularize the domain-specific features.

### 3.1. Conditional Bilateral Style Transfer (CBST)

According to the Underwater Model mentioned in [9], the formation of underwater images can be modeled as:

$$I_\lambda(x) = J_\lambda(x) \cdot t_\lambda(x) + (1 - t_\lambda(x)) \cdot B_\lambda, \quad \lambda \in \{red, green, blue\}, \tag{1}$$

where $x$ is a point in the underwater scene, $I_\lambda(x)$ is the image captured by the camera, $J_\lambda(x)$ is the clear latent image at point $x$, $t_\lambda(x)$ is the residual energy ratio of $J_\lambda(x)$ after reflecting from point $x$ in the underwater scene and reaching the camera, $B_\lambda$ is the homogeneous background light, and $\lambda$ is the light wavelength. According to the equation mentioned above, it can be concluded that an underwater image $I$ is linearly transformed from the clear latent image $J$ on color space. In other words, if two underwater images have the same clear latent image, they can be linearly transformed from each other. Therefore, Conditional Bilateral Style Transfer is proposed to synthesize underwater images. It learns a local affine color transformation from a low-resolution style image and applies the transformation to the content image. CBST is designed based on the state-of-the-art photorealistic style transfer model BST [56]. Compared to BST, CBST makes two major promotions. First, arbitrary style transfer is not necessary, all AdaIN module in the original network is replaced by Conditional Instance Normalization (CIN) [21]. CIN can be represented as:

$$CIN(x; s) = \gamma^s\left(\frac{x - \mu(x)}{\sigma(x)}\right) + \beta^s, \tag{2}$$

where $\mu$ and $\sigma$ are $x$'s mean and standard deviation taken across spatial axes, and $\gamma^s$ and $\beta^s$ are scale and shift parameters corresponding to target style $s$. The model with CIN is much easier to train, consuming less time in the inference stage, because there is no need to extract style images' features from VGG. Second, inspired by the idea of region loss [54], Mask loss is introduced. Style transfer may make the image too stylized, leading to a change of contents in the image. However, if the detector is trained on stylized images with essential content changes, the mismatch of annotation will result in a great performance decline.

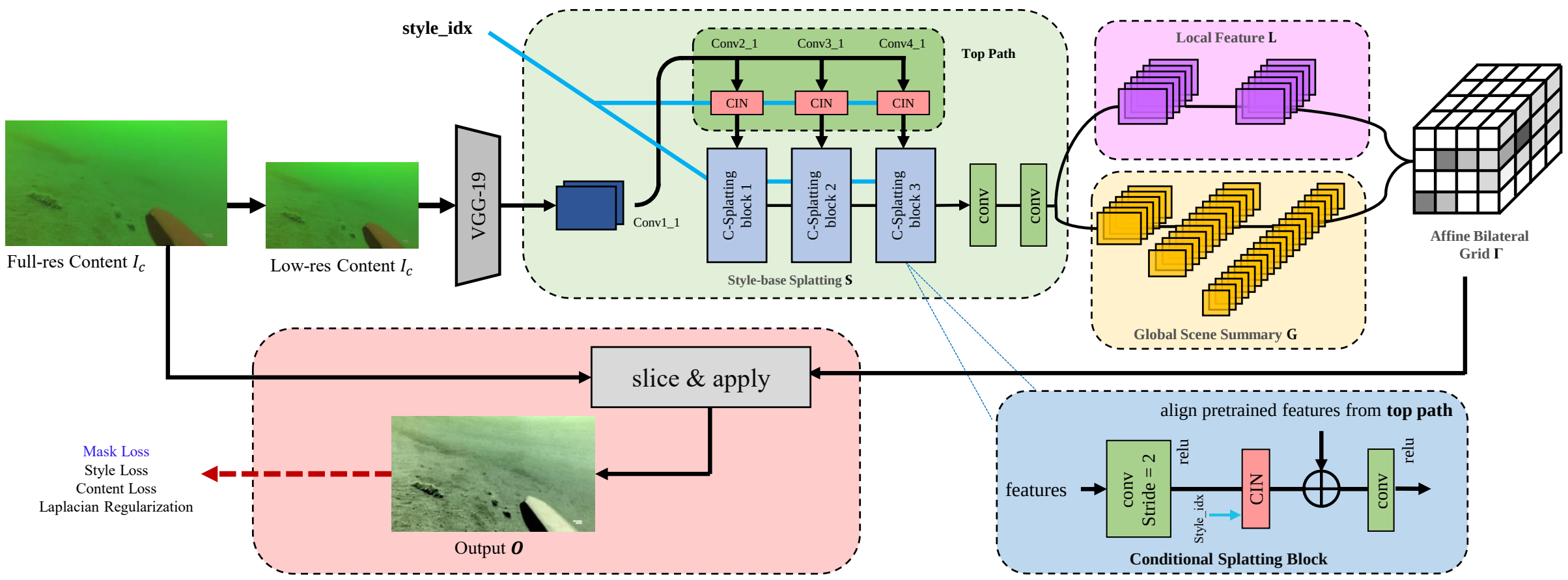

**Figure 4:** Framework of Conditional Bilateral Style Transfer. The basic framework is modified from BST [56]. Compared with BST, the AdaIN modules of top path and Splatting Blocks are replaced by CIN. What's more, Mask loss is added in the training stage for protecting important semantic contents.

The architecture of CBST is shown in Fig. 4. First, a low resolution content image $\boldsymbol{C}_{low}$ is sent into VGG19 to extract features $\boldsymbol{F}_{conv1_1}$, $\boldsymbol{F}_{conv2_1}$, $\boldsymbol{F}_{conv3_1}$ and $\boldsymbol{F}_{conv4_1}$.

Second, $\boldsymbol{F}_{conv1_1}$ is sent into three successive splatting blocks. Each splatting block contains a stride-2 convolutional layer, a CIN and a stride-1 convolutional layer. What's more, $\boldsymbol{F}_{conv2_1}$, $\boldsymbol{F}_{conv3_1}$ and $\boldsymbol{F}_{conv4_1}$ after CIN are added to the top path of three splatting block respectively. After three splatting blocks, two extra convolutional layers are followed.

Third, the network is split into two asymmetric paths. A fully-convolutional local path that learns local color transforms and thereby sets the grid resolution. And a global path with both convolutional and fully connected layers considers all the pixels of features to learn a summary of the scene and helps spatially regularize the transforms. Local Feature and Global Scene Summary are combined to obtain a color transformation $\boldsymbol{A} \in \mathbb{R}^{16\times16\times96}$. $\boldsymbol{A}$ can be viewed as a $16 \times 16 \times 8$ bilateral grid, where each grid cell contains 12 numbers, one for each coefficient of a $3 \times 4$ affine color transformation matrix, which means $\boldsymbol{A}_{i,j,k} \in \mathbb{R}^{3\times4}$.

Fourth, a full-resolution content image $\boldsymbol{C}$ is fed into the Guidance Map Auxiliary Network to obtain guidance map $\boldsymbol{g} \in \mathbb{R}^{h\times w}$, and the bilateral grid is upsampled using the guidance map $\boldsymbol{g}$,

$$\overline{\boldsymbol{A}}_{x,y} = \sum_{i,j,k} \tau(s_x x - i)\tau(s_y y - j)\tau(d \cdot g_{x,y} - k)\boldsymbol{A}_{i,j,k}, \quad (3)$$

where $\boldsymbol{C} \in \mathbb{R}^{h\times w\times3}$, and $h$ and $w$ are the height and width of $C$. $\overline{\boldsymbol{A}}_{x,y} \in \mathbb{R}^{3\times4}$, and $(x, y)$ is the pixel coordinates of the content image. $s_x$ and $s_y$ are the width and height ratios of the grid's dimensions w.r.t the full-resolution image's dimensions, and $\tau(\cdot) = max(1 - |\cdot|, 0)$.

Fifth, An affine transform is applied to the content image. $\overline{\boldsymbol{A}}_{x,y}$ can be split into a color mapping matrix $\overline{\boldsymbol{Am}}_{x,y} \in \mathbb{R}^{3\times3}$ and a color bias $\overline{\boldsymbol{Ab}}_{x,y} \in \mathbb{R}^{3\times1}$, the color affine transform can be written as:

$$\boldsymbol{O}_{x,y} = \boldsymbol{C}_{x,y} \otimes \overline{\boldsymbol{Am}}_{x,y} + \overline{\boldsymbol{Ab}}_{x,y}, \quad (4)$$

$\boldsymbol{O} \in \mathbb{R}^{h\times w\times3}$ is the output image, $\boldsymbol{O}_{x,y} \in \mathbb{R}^{3\times1}$, and $\otimes$ means matrix multiplication.

As for the loss function of CBST, Mask loss is proposed to prevent the change of important semantic information:

$$L_{mask} = \frac{1}{h \cdot w \cdot 3}||(\boldsymbol{O} - \boldsymbol{C}) \cdot \boldsymbol{M}||, \quad (5)$$

$$\mathbf{M}(i,j) = \begin{cases} 1, & \text{if } (i,j) \in \mathbb{B}^2 \quad (6a) \\ 0.01, & \text{otherwise} \quad (6b) \end{cases}$$

where $\mathbb{B}^2$ denotes the set of pixels in the boxes of the objects, $\boldsymbol{M} \in \mathbb{R}^{h x w x 1}$ denotes the mask created by bounding boxes of objects.

In general, L1 loss and content loss [20] are used to supervise the content of the output images. L1 loss directly minimizes the distance between content images and output images, which forces the output images to be the same as the content images, which will easily lead to a collapse in style learning. Content loss minimizes the distance on the VGG-extracted features in order to obtain semantic consistency. However, the VGG has not been trained to recognize underwater objects in the proposed dataset. Thus, feature consistency does not guarantee semantic consistency. For example, in Fig. 7 (in the original manuscript), BST generates echinus with different appearances from content images, but it has the same VGG features. Compared with L1 loss, Mask loss focuses on minimizing the area where objects exist and only uses a small constraint on backgrounds, which avoids the collapse of style learning. Compared with content loss, Mask loss is applied to the outputs, providing more accurate supervision to protect the semantic contents.

The total loss function of CBST is:

$$L = \lambda_c L_c + \lambda_{sa} L_{sa} + \lambda_r L_r + \lambda_{mask} L_{mask}, \quad (7)$$

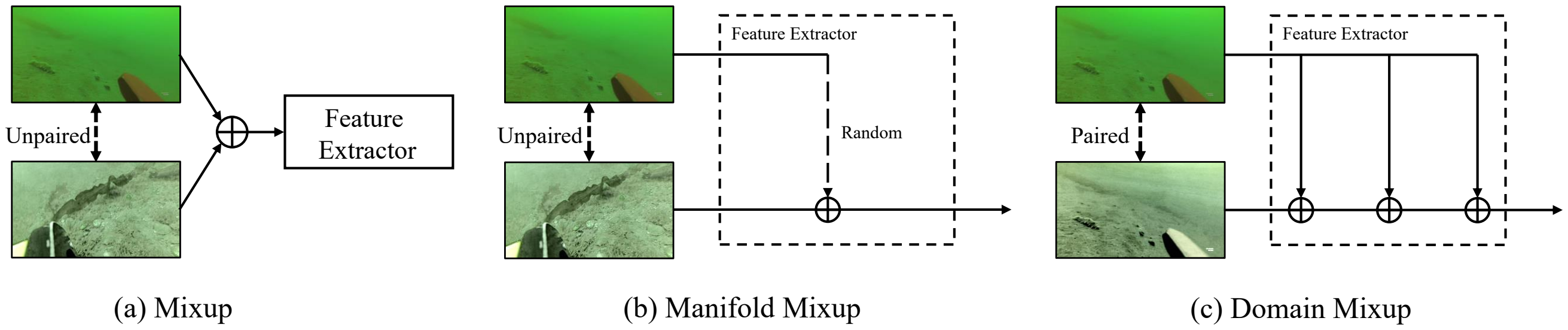

**Figure 5:** The illustration of different Mixup operations. (a) Original Mixup performs interpolation on the image level between two unpaired images. (b) Manifold Mixup performs interpolation on a random feature layer between two unpaired images. (c) The proposed Domain Mixup performs interpolation on a multiple feature layer between two paired images which are generated from style transfer.

where $L_c$ is the content loss, $L_{sa}$ is the style loss and $L_r$ is the bilateral-space Laplacian regularizer, which are the same as BST [56]. We set hyper-parameters $\lambda_c = 0.5, \lambda_{sa} = 1, \lambda_r = 0.015, \lambda_{mask} = 1$ in our experiments.

### 3.2. Domain Mixup (DMX)

CBST can realize the data augmentation on the image level, but the domain diversity is still limited in the source domains provided by the dataset. In order to further enrich the domain diversity, more domain diversity outside the dataset can be obtained by sampling on the features level. Through CBST, the original image $I_1$ and its corresponding generated image $I_2$ are obtained. In the backbone, the latent features of $I_1$ and $I_2$ contain domain information and semantic information. Since the ground truth of $I_1$ and $I_2$ are the same, their latent semantic information is the same, while the irrelevant domain information is different. If the latent features of $I_1$ and $I_2$ are interpolated, the semantic information is unchanged, and the domain information is interpolated. Since the domain manifold is more flattened in the latent space than in the input space, linear interpolation can generate new domains inside the domain convex hull.

In detail, assuming that the features contain two parts: domain-invariant features and domain-specific features, which are depicted as:

$$F = F_d + F_s, \tag{8}$$

where $F_d$ denotes domain-invariant features, and $F_s$ denotes domain-specific features. $F, F_d, F_s \in \mathbb{R}^{H\times W\times C}$. When the underwater environment changes (light changes, water quality changes, etc.), only $F_d$ changes while $F_s$ maintains the same. Thus, in the ideal condition, $F_s$ is enough for detectors to make correct predictions. When we perform Mixup on two pair features:

$$\lambda * F_1 + (1-\lambda) * F_2 \tag{9}$$
$$=(\lambda * F_{1,d} + (1-\lambda) * F_{2,d}) + F_s, \tag{10}$$

where $F_{1,d}$ and $F_{2,d}$ denotes the domain-specific feature of two images. And two paired features have the same domain-invariant features. Thus, the semantic information is unchanged, and the domain information is interpolated.

In the implementation, consider that $K = [k_1, k_2, ..., k_n]$ are the selected layers of the backbone to be performed Mixup.

$I_1$ and $I_2$ are fed into the backbone simultaneously (Main Stream and Assistance Stream). Two streams share the same parameters. The features in Assistance Stream are used to augment the features of Main Stream. In Main Stream, the k-th layer latent features can be expressed as:

$$h_{1,k} = \begin{cases} \lambda_k \cdot f_k(h_{1,k-1}) \\ +(1-\lambda_k)\cdot f_k(h_{2,k-1}), & k \in K, \quad \text{(11a)} \\ f_k(h_{1,k-1}), & \text{otherwise}, \quad \text{(11b)} \end{cases}$$

while in Assistance Stream, the latent features can be expressed as:

$$h_{2,k} = f_k(h_{2,k-1}), \tag{12}$$

where $h_{1,k}$ and $h_{2,k}$ are the feature maps of k-th layer of Main Stream and Assistance Stream respectively, and $\lambda_k \sim Beta(\alpha, \alpha)$ is the Mixup ratio of k-th layer, for $\alpha \in (0, \infty)$. The gradient is backpropagated through both Main Stream and Assistance Stream. The features after Domain Mixup ($h_{1,k}$ in k-th layer) are used in the Main Stream and fed into the detection head.

Although the proposed Domain Mixup is inspired by Mixup, these two methods are quite different. Mixup [63] performs interpolation on the image level between two unpaired images (Fig. 5 (a)). The ground truths of the mixed image will be the combination of two unpaired images. The proposed Domain Mixup performs interpolation on the feature level between two paired images whose ground truths are the same (Fig. 5 (c)). Besides, Manifold Mixup [48] also performs interpolation on the feature level, but it randomly selects one layer to interpolate, and unpaired images are leveraged (Fig. 5 (b)). What's more, DM-ADA [58] also proposes a Domain Mixup method, performing Mixup on the image level and feature level at the same time. However, the Domain Mixup in DM-ADA also uses unpaired images. DM-ADA was specially designed for the recognition task, and can not be applied to the detection task.

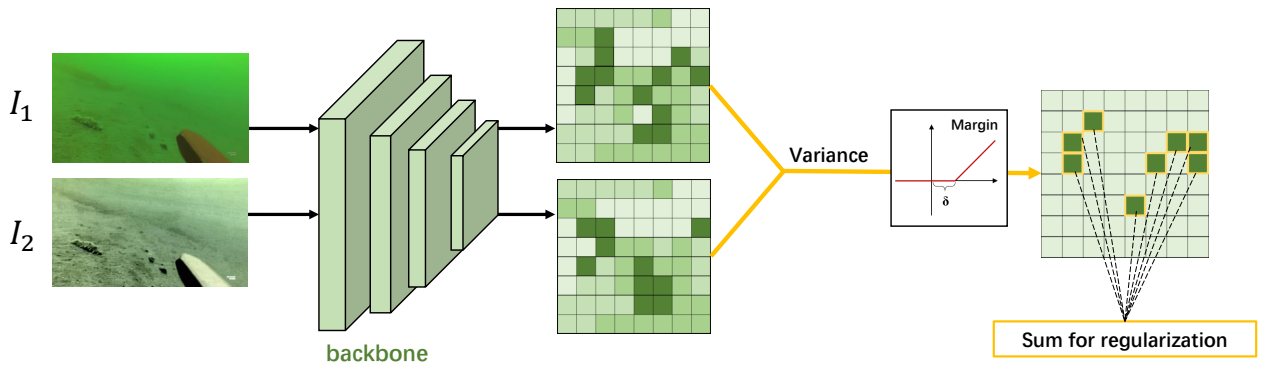

**Figure 6:** The illustration of SSMC loss. Paired images are fed into backbone to obtain paired features. Variance is caculated on the paired images. After margin operation, the highest k value will be summed for regularization.

Here is the discussion about why DMX works. $\mathcal{Z}$ is used to represent the latent features of all layers. It can be assumed that the data distribution $\mathcal{Z}$ has two independent prior distributions: the domain distribution $\mathcal{D}$ and the semantic contents distribution $\mathcal{S}$ (see Fig. 1). $\mathcal{D}$ constructs a latent domain manifold. From the perspective of the Bayesian model, we can get:

$$\begin{aligned} p(S \mid Z) &= \int p(S, D \mid Z) dD \\ &= \int \frac{p(Z \mid D, S) \cdot p(D) \cdot p(S)}{p(Z)} dD \\ &= E_D \left[ \frac{p(Z \mid D, S) \cdot p(S)}{p(Z)} \right], \end{aligned} \tag{13}$$

The detector can be represented as $q(S|Z, \theta)$ with learnable parameters $\theta$. The goal of the detector is to reduce the divergence between distribution $p(S|Z)$ and $q(S|Z, \theta)$. Leveraging Equation (4), the optimization goal is:

$$\begin{aligned} & argmin \;\; KL(p(S|Z)||q(S|Z,\theta)) \\ &= argmin \;\; \int p(S|Z) \cdot log \frac{p(S|Z)}{q(S|Z,\theta)} dS \\ &= argmax \;\; E_D[E_S[p(Z|D,S) \cdot log\; q(S|Z,\theta)]], \end{aligned} \tag{14}$$

In the discrete condition, $p(Z|D, S)$ can be approximated with empirical data distribution $\frac{1}{N_{D,S}} \sum_{n=1}^{N_{D,S}} \delta_{Z_n}(Z)$ ($N_{D,S}$ is the number of data with certain $D$ and $S$), and approximate the expectation in the optimization goal (14) by sampling on the domain distribution. The source domains of S-UODAC2020 are limited, occupying 6 small areas of the whole domain manifold. Without a large number of samples, the approximations are inaccurate. As a result, the model probably "memorizes" these domains. A solution to this problem is to sample as many domains as possible during training. CBST generates images of different domains with the same semantic contents, which can be seen as sampling on $D$ given $S$. Assumed source domains construct a convex hull on the domain manifold, DMX can sample more domains inside the domain convex hull by interpolation.

### 3.3. Spatial Selective Marginal Contrastive Loss

Given two images with the same semantic contents in different domains, it can be assumed that the hidden features extracted by the backbone are the same. We leverage the ideas of contrastive learning to design Spatial Selective Marginal Contrastive Loss (SSMC loss). In detail, given the original image $I_1$ and its corresponding image $I_2$, Spatial Contrastive Loss (SC Loss) can be computed as:

$$L_{SC} = ||F(I_1) - F(I_2)||_2^2, \tag{15}$$

where $F$ denotes the backbone, $F(I_1), F(I_2) \in \mathbb{R}^{H \times W \times C}$, $H, W, C$ are the height, width and channels of the features, and $|| \cdot ||$ denotes L2norm. However, too constrained regularization may adversely affect the discriminative capacity of the detector. In order to address the issue, two solutions are proposed. First, the pixels with the greater changes are selected for regularization. Therefore, Spatial Selective Contrastive Loss (SSC Loss) is proposed:

$$V = (F(I_1) - F(I_2))^2, \tag{16}$$

$$L_{SSC} = kMaxpooling(\frac{1}{C} \sum_{j}^{C} V_j), \tag{17}$$

where $V$ denotes the variance matrix, $V_j$ is the j-th channel of $V$, and $kMaxpooling$ is defined as:

$$kMaxpooling(H) = \frac{1}{k} \sum_{i}^{k} topk(H), \tag{18}$$

where $topk$ is the highest k value in $H$, $H \in \mathbb{R}^{H \times W}$, $k = (H \times W/16)$. Second, Spatial Selective Marginal Contrastive Loss (SSMC loss) is designed so that the expected value of the spatial variance lies within a specified margin $\delta$ rather than being close to zero,

$$L_{SSMC} = kMaxpooling(max(\frac{1}{C} \sum_{j}^{C} V_j - \delta, 0)), \tag{19}$$

The proposed SSMC loss can regularize the domain-specific pixels, but allows room to keep the discriminative power of the network. The pipeline of SSMC is shown in Fig. 6.

In conclusion, the total loss function of our proposed method can be computed as:

$$L_{total} = \lambda_{SSMC} * L_{SSMC} + L_{rpn_cls} + L_{rpn_reg} + L_{cls} + L_{reg}, \tag{20}$$

where $L_{rpn_cls}$ and $L_{rpn_reg}$ denote the classification loss and the regression loss of the RPN, and $L_{cls}$ and $L_{reg}$ denote the classification loss and the regression loss of the R-CNN head. $\lambda_{SSMC}$ is the balanced coefficicent of SSMC loss.

## 4. Experiments and Discussion

We give an overview of how this section is organized. First, the underwater dataset is introduced where experiments are implemented. Second, our method is compared with other domain generalization methods and achieves better performance. Third, comprehensive ablation studies are designed to prove the effectiveness of all the modules of DMCL and find out the best hyper-parameters, and experiments on fewer source domains are to test the robustness of DMCL. Fourth, to prove that CBST is a better choice for our method, a comparison of image quality, detection performance, and inference speed is conducted with other real-time style transfer methods. Fifth, analysis of the statistics of the features and the visualization of t-SNE [39] demonstrate that our method helps the model learns domain-invariant features. Sixth, a comparison of other domain generalization datasets further proves the effectiveness of our method.

**Table 1**
The Performance of Different Domain Generalization Methods. Training Dataset is the Source Domains Dataset of S-UODAC2020. "Ave." Denotes the mAP@50 Performance on the Target Domain of S-UODAC2020, Representing the Domain Generalization Capacity of the Model.

| Detector | Method | Input size | Backbone | Epochs | Echinus (%) | Starfish (%) | Holothurian (%) | Scallop (%) | Ave. (%) |
|---|---|---|---|---|---|---|---|---|---|
| YOLOv3 | DeepAll | DarkNet53 | 300 | $416 \times 416$ | 70.28 | 31.83 | 27.67 | 41.74 | 42.88 |
| | DANN [14] | DarkNet53 | 300 | $416 \times 416$ | 63.67 | 24.01 | 25.78 | 30.28 | 37.32 |
| | CrossGrad [44] | DarkNet53 | 300 | $416 \times 416$ | 71.21 | 31.41 | 32.17 | 32.46 | 41.81 |
| | JiGEN [3] | DarkNet53 | 300 | $513 \times 513$ | 73.15 | **38.56** | 34.57 | 35.06 | 45.34 |
| | DG-YOLO [35] | DarkNet53 | 300 | $416 \times 416$ | 62.74 | 26.83 | 32.84 | 34.54 | 39.24 |
| | **DMCL(Ours)** | DarkNet53 | 300 | $416 \times 416$ | **73.26** | 36.69 | **43.79** | **59.61** | **53.34** |
| Faster R-CNN | DeepAll | ResNet50 | 12 | $1333 \times 800$ | 74.79 | 36.59 | 43.12 | 40.94 | 48.86 |
| | Mixup [63] | ResNet50 | 12 | $640 \times 640$ | 70.23 | 34.58 | 40.01 | 18.86 | 40.92 |
| | DANN [14] | ResNet50 | 12 | $1333 \times 800$ | **78.62** | 42.76 | 50.60 | 43.48 | 53.87 |
| | DANN [14] | ResNet101 | 24 | $1333 \times 800$ | 73.23 | 49.92 | 50.96 | 50.61 | 56.18 |
| | CCSA [40] | ResNet50 | 12 | $1333 \times 800$ | 76.71 | 36.85 | 40.58 | 37.46 | 47.90 |
| | CrossGrad [44] | ResNet50 | 12 | $1333 \times 800$ | 77.67 | 45.43 | 49.80 | 42.40 | 53.83 |
| | MMD-AAE [25] | ResNet50 | 12 | $1333 \times 800$ | 75.73 | 35.00 | 43.31 | 44.86 | 49.73 |
| | CIDDG [32] | ResNet50 | 12 | $1333 \times 800$ | 76.37 | 39.89 | 42.27 | 43.65 | 50.55 |
| | CIDDG [32] | ResNet101 | 24 | $1333 \times 800$ | 74.04 | 48.98 | 49.71 | 45.67 | 54.60 |
| | JiGEN [3] | ResNet50 | 12 | $768 \times 768$ | 76.15 | 39.06 | 50.27 | 41.44 | 51.73 |
| | JiGEN [3] | ResNet101 | 24 | $768 \times 768$ | 75.92 | 47.01 | 51.37 | 46.50 | 55.20 |
| | **DMCL (Ours)** | ResNet50 | 12 | $512 \times 512$ | 78.44 | **54.62** | **53.15** | **59.23** | **61.36** |
| RoIAttn [33] | | ResNet50 | 12 | $1333 \times 800$ | 74.41 | 43.28 | 50.01 | 42.66 | 52.59 |
| VFNet [64] | | ResNet50 | 12 | $1333 \times 800$ | 72.97 | 43.21 | 44.02 | 47.84 | 52.01 |

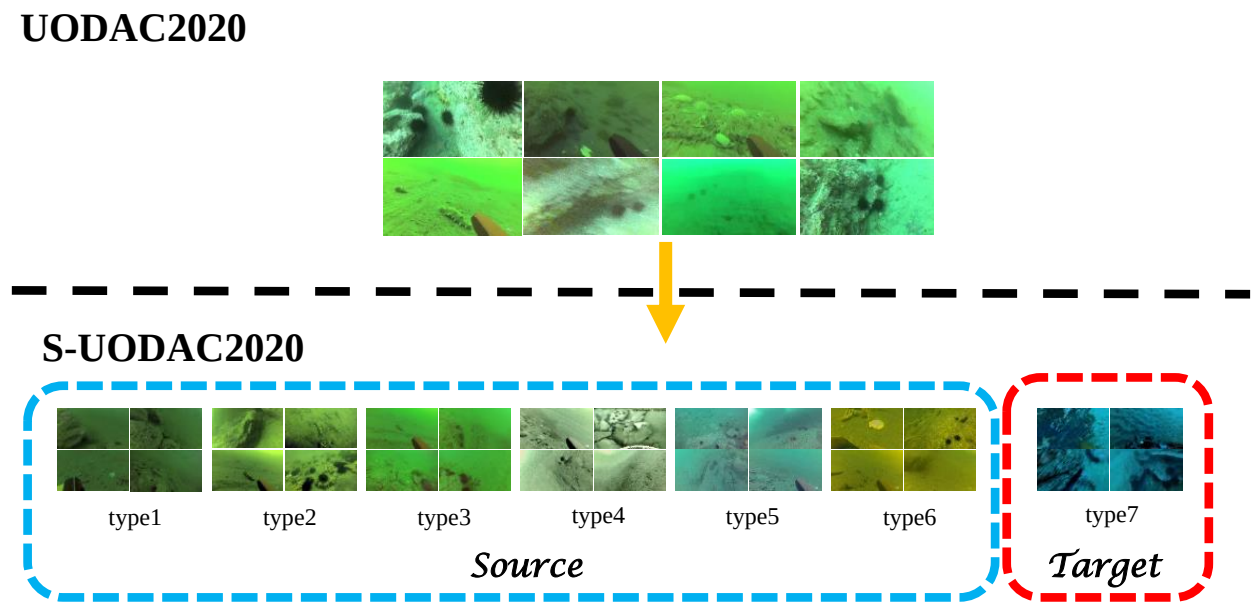

**Figure 7:** Illustration of S-UODAC2020 dataset. UODAC2020 is randomly split into 7 sets on average, and each set contains 791 images, 5537 images in total. These 7 sets are transferred to 7 domains respectively (*type1* to *type7*, *type1* to *type6* are source domains $\mathcal{S}$ and *type7* is target domain $\mathcal{T}$).

## 4.1. Problem Statement and the Proposed Dataset

In [35], a domain generalization dataset S-URPC2019 for underwater detection is proposed. However, there is a very strong assumption in the S-URPC2019 dataset that each domain is composed of semantically identical images in different domains. This assumption is highly unrealistic in practice. Therefore, we aim to leverage the idea of the generation of S-URPC2019 but fix the unrealistic assumption. We proposed Synthetic Underwater Object Detection Algorithm Contest 2020 (S-UODAC2020) dataset, which is generated from Underwater Object Detection Algorithm Contest 2020 (UODAC2020)[2]. Different from the dataset processing of S-URPC2019, UODAC2020 is randomly split into 7 sets on average, and each set contains 791 images. We transfer 7 sets to 7 domains (*type1* to *type7*) respectively by style transfer model $WCT^2$ [62], in which *type1* to *type6* are source domains $\mathcal{S}$ and *type7* is target domain $\mathcal{T}$ (Fig. 7). $WCT^2$ is an arbitrary style transfer and requires the style images. The style images of *type1* to *type5* are selected from UODAC2020 dataset. The style images of *type6* and *type7* are from the internet, and they are quite different from *type1* to *type5*. For source data $x_s \in \mathcal{S}$, class label $c$ and the bounding box coordinates $b = [x, y, w, h]$ are available. Models should be trained on source domains and evaluated on the target domain. The target domain data are totally not accessible during training. S-UODAC2020 contains four classes: echinus, holothurian, starfish, and scallop.

## 4.2. Experiments in Domain Generalization Setting

### 4.2.1. Experimental setup

DMCL are applied to YOLOv3 and Faster R-CNN with FPN, which are widely used one-stage detectors and two-stage detectors, respectively. Both models are trained on an Nvidia GTX 1080Ti GPU with PyTorch implementation. Models are trained on S-UODAC2020 training set (*type1-6*) and evaluated on S-UODAC2020 test set (*type7*). For YOLOv3, we train it for 100 epochs with batch size 8. Adam is employed for optimization with learning rate set to 0.001, $\beta_1 = 0.9$, and $\beta_2 = 0.999$. Multi-scale training is used. IoU, confidence, and NMS thresholds are set to 0.5, 0.02, and 0.5, respectively. Accumulating gradients (every 2 iterations) are applied during training. Mixup is performed on the features from layers 36, 46, 55, and 71 of DarkNet. SSMC loss is applied on layer 71 of DarkNet. For Faster R-CNN, we train it for 24 epochs with batch size 4. SGD is employed for optimization with learning rate, momentum, and weight decay set to 0.02, 0.9, and 0.0001, respectively. IoU, confidence, and NMS thresholds are set to 0.5, 0.05, and 0.5, respectively. Mixup is performed on the features of the last three stages of ResNet-50. SSMC loss is applied in the last stage of ResNet. None of any other data augmentation methods except horizontal-flip is used unless we mention it.

[2] http://uodac.pcl.ac.cn/

**Table 2**
The Training Cost and Model Complexity of Different Domain Generalization Methods. FPS is tested in Nvidia 1080Ti GPU.

| Method | Input size | Training Time (min) | Memory (M) | GFlops | FPS | Params (M) |
|---|---|---|---|---|---|---|
| DeepAll | $1333 \times 800$ | 196 | 3784 | 216.31 | 11.6 | 41.14 |
| Mixup [63] | $640 \times 640$ | **100** | 3018 | 91.02 | 18.7 | 41.14 |
| DANN [14] | $1333 \times 800$ | 202 | 3804 | 216.31 | 11.6 | 41.14 |
| CCSA [40] | $1333 \times 800$ | 204 | 3785 | 216.31 | 11.6 | 41.14 |
| CrossGrad [44] | $1333 \times 800$ | 735 | 8504 | 216.31 | 11.6 | 41.14 |
| MMD-AAE [25] | $1333 \times 800$ | 437 | 6680 | 216.31 | 11.6 | 41.14 |
| CIDDG [32] | $1333 \times 800$ | 204 | 3885 | 216.31 | 11.6 | 41.14 |
| JiGEN [3] | $768 \times 768$ | 162 | 2725 | 124.94 | 16.3 | 41.14 |
| **DMCL (Ours)** | $512 \times 512$ | 204 | **1814** | **63.26** | **21.8** | 41.14 |

**Table 3**
Ablation Study. “DMX*" Means Mixup on Feature Level in a Randomly Chosen Layer. “Output Before Mixup" Means Using the Features Before Mixup for Detection. “Detach Mixup" Means Backpropagating Gradient Through Only Main Stream. “Ours w. SC Loss" and “Ours w. SSC Loss" Means Replacing SSMC loss with SC Loss and SSC Loss Respectively.

| Method | Echinus(%) | Starfish(%) | Holothurian(%) | Scallop(%) | Ave.(%) |
|---|---|---|---|---|---|
| DeepAll | 74.79 | 36.59 | 43.12 | 40.94 | 48.86 |
| Only CBST | 76.88 | 49.09 | 50.49 | 56.23 | 58.17 |
| CBST + DMX_IN | 75.56 | 47.72 | 46.43 | 47.19 | 54.23 |
| CBST + DMX* | 76.79 | 51.96 | 51.26 | 57.20 | 59.30 |
| CBST + DMX | 77.11 | 54.73 | 51.70 | 57.74 | 60.32 |
| Ouput Before Mixup | 76.24 | 50.37 | 51.31 | 56.74 | 58.67 |
| Detach Mixup | 76.49 | **55.10** | 51.70 | 57.74 | 60.26 |
| DMCL w. SC Loss | **78.85** | 47.53 | **54.83** | 56.66 | 59.47 |
| DMCL w. SSC Loss | 78.18 | 54.53 | 51.74 | **59.43** | 60.97 |
| **DMCL** | 78.44 | 54.62 | 53.15 | 59.23 | **61.36** |

**Table 4**
Ablation Study of SSMC Loss coefficient.

| $\lambda_{SSMC}$ | 0 | 1 | 5 | 10 | 15 |
|---|---|---|---|---|---|
| mAP(%) | 60.32 | 60.83 | 60.91 | 61.36 | 59.27 |

#### 4.2.2. *Comparison with other domain generalization methods*

Since there are few domain generalization methods for object detection, some of domain adaptation methods and domain generalization methods for recognition are chosen and modified to fit for domain generalization of object detection. The following methods are compared with the proposed method.

- **DeepAll** follows the traditional training paradigm, which is simply training the detector on all source domains.
- **Mixup** [63] is a data augmentation method which is originally designed for the recognition task. [65] applies it on object detection task, which is leveraged in this experiment.
- **DANN** [14] is a domain adaptation method that uses a domain classifier to distinguish data from source and target for adversarial training. It is modified to distinguish data between source domains with a single domain classifier. In Faster R-CNN, the features of the 2nd stage of ResNet are chosen for adversarial training, while in YOLOv3, the features of the 36th layer of DarkNet are chosen.
- **CCSA** [40] designs the Contrastive Semantic Alignment (CSA) loss to align the features of source domains. The CSA loss is specially designed for the recognition task (it uses the class information), and it can be easily applied to the R-CNN head in Faster

**Table 5**
Ablation Study of Mixup Stage.

| row_idx | Stage1 | Stage2 | Stage3 | Stage4 | Echinus(%) | Starfish(%) | Holothurian(%) | Scallop(%) | Ave.(%) |
|---|---|---|---|---|---|---|---|---|---|
| 1 | ✓ | ✓ | ✓ | ✓ | 77.06 | 54.16 | 52.08 | 57.85 | 60.29 |
| 2 | ✓ | ✓ | ✓ | | 77.38 | 50.45 | **52.82** | 54.76 | 58.85 |
| 3 | | ✓ | ✓ | ✓ | 77.11 | 54.73 | 51.70 | 57.74 | **60.32** |
| 4 | ✓ | ✓ | | | 76.66 | 52.62 | 51.59 | **58.73** | 59.90 |
| 5 | | ✓ | ✓ | | **77.69** | **55.49** | 50.37 | 57.26 | 60.20 |
| 6 | | | ✓ | ✓ | 76.20 | 52.53 | 51.41 | 58.59 | 59.68 |
| 7 | ✓ | | | | 75.75 | 53.52 | 49.60 | 54.93 | 58.45 |
| 8 | | ✓ | | | 76.74 | 54.50 | 49.03 | 57.40 | 59.42 |
| 9 | | | ✓ | | 76.17 | 52.57 | 50.74 | 56.60 | 59.02 |
| 10 | | | | ✓ | 76.52 | 54.75 | 51.18 | 57.04 | 59.87 |

R-CNN, because the RPN and RoI Align decouple the classification and regression. However, it is not suitable for YOLOv3. The negative proposals (background) do not participate in the CSA loss calculation.

- **CrossGrad** [44] follows the idea of defense against adversarial attack, and the loss of recognition is replaced with loss of detection. Domain classifier is set the same as DANN.
- **MMD-AAE** [25] aligns the features of source domains by adversarial learning and MMD loss. It is only implemented on Faster R-CNN. The shared fc layer in the R-CNN head is regarded as the encoder with 1024 hidden neurons, and the classification fc layer in the R-CNN head is the classifier. The decoder and discriminator with 2 fc layers is added to the R-CNN head.
- **CIDDG** [32] uses class-conditioned and class prior-normalized domain classifier to perform domain adversarial training. It is only implemented on Faster R-CNN for the same reason as CCSA. All of the domain classifiers are applied to the features of the shared fc layer in the R-CNN head. The negative proposals do not participate in the adversarial training.
- **JiGEN** [3] designs an auxiliary jigsaw puzzle task to increase the domain generalization capacity. In Faster R-CNN, the features of the last stage of ResNet are chosen for jigsaw puzzle task classification, while in YOLOv3, the features of 36th layer of DarkNet are chosen. Input sizes are $513 \times 513$ and $768 \times 768$ in YOLOv3 and Faster R-CNN, respectively, for breaking up the puzzle.
- **DG-YOLO** [35] leverages adversarial training and IRM penalty to align features of source domains. It is re-trained and evaluated on S-UODAC2020 dataset.

CCSA, MMD-AAE, and CIDDG are only implemented on Faster R-CNN in the R-CNN head, where the detection task becomes instance classification. In Table 1, It can be concluded that our method achieves the highest performance in both YOLOv3 (53.34%) and Faster R-CNN (61.36%), which is higher than DANN in YOLOv3 (45.34%) and Faster R-CNN (53.87%).

**Table 6**
The Performance When It Comes to Fewer Source Domains. There are Three Groups of Comparison, Training on *type1-6*, *type1-4* and *type1-2*. All Models are Evaluated on *type7*.

| Source | Method | mAP on type8 |
|---|---|---|
| 1-6 | DeepAll | 48.86 |
| 1-6 | DANN | 53.87 |
| 1-6 | **Ours** | 61.36 |
| 1-4 | DeepAll | 49.54 |
| 1-4 | DANN | 49.67 |
| 1-4 | **Ours** | 52.95 |
| 1-2 | DeepAll | 42.58 |
| 1-2 | DANN | 41.26 |
| 1-2 | **Ours** | 43.25 |

We also use a larger backbone (ResNet101) on DANN, CIDDG and JiGEN. Although using ResNet101 brings higher performance to DANN (56.18%), CIDDG (54.60%) and JiGEN (55.20%), they still can not achieve the same level as the proposed method with ResNet50.

The performance of the original mixup is even lower than DeepAll (40.92% vs 48.86%). Because underwater images are blurred and distorted, the objects in the images are vague. Mixup performed on unpaired images leads to semantic confusion, making the detector insufficient to learn from data. Unlike the original mixup, our DMX operation is performed on paired images, which maintains semantic consistency. Also, it can be observed that domain generalization methods that are applied on the image level (DMCL, DANN, CrossGrad, and JiGEN) are generally higher than those which work on the instance level (CCSA, MMD-AAE, and CIDDG). It implies that the methods which perform well in the recognition task are not entirely suitable for the detection task.

We also compared our methods with two recent object detection methods: an underwater object detector RoIAttn [33] and a generic object detector VFNet [64] in Table 1.

**Table 7**
Speed Comparison (in FPS) for 512 × 512 Images and Performance Comparison When Being Used for Online Data Augmentation. Speeds are Obtained With an Nvidia 1080Ti GPU and Averaged Over 100 Images.

| Method | MST [21] | AdaIN [20] | BST [56] | CBST |
|---|---|---|---|---|
| FPS | 39.98 | 15.63 | 78.62 | **105.71** |
| mAP | 55.71 | 51.91 | 56.00 | **58.17** |

RoIAttn and VFNet obtain 52.59% and 52.01% mAP, respectively. Although the recent detectors achieve competitive performances due to their strong extraction capabilities, they are still inferior to our methods because of the vulnerability to domain shift.

The comparisons of the training cost and model complexity are shown in Table 2. For a fair comparison, all the methods are implemented in Faster R-CNN+FPN+R50 framework. All the DG methods compared in Table 1 only change the training paradigm of the Faster R-CNN and YOLO, and do not modify their original structure. As a result, their params are the same. Besides, their Gflops and FPS is related to their input size. As for the training cost, DMCL has a similar training time to DeepAll, DANN, CCSA, and CIDDG. What's more, DMCL spends the minimum memory compared with other methods.

### *4.2.3. Ablation study*

The results of the ablation study are shown in Table 3. If source images and images generated by CBST are randomly chosen for training ("Only CBST"), the performance increases drastically to 58.17% compared with DeepAll (48.86%). Moreover, adding DMX ("CBST+DMX") further increases the performance to 60.32%, while performing Domain Mixup on input level ("DMX_IN") reduces the performance to 54.23%. This proves that the latent feature space is more flattened, so it is more reasonable to synthesize new domains by interpolating on the feature level. Moreover, Mixup on feature level in a randomly chosen layer [48] ("CBST+DMX*") gets 59.30% mAP, and can not achieve the same performance as "CBST+DMX". Using the features before Mixup for detection ("Output Before Mixup") has little increase in performance (58.67%) compared with "CBST+DMX". There is another backpropagation way, which is backpropagating gradient through only Main Stream ("Detach Mixup"). The performance of "Detach Mixup" is almost equal to "CBST+DMX", which is 60.26%.

Full DMCL achieves a performance of 61.36%. If SSMC loss is replaced with SC Loss, performance will degrade (59.47%), for the constraint is too strict. Compared with SC Loss, SSC Loss relaxes the constraint, so the performance increases to 60.97%. However, in the later stage of training, features from different domains are similar enough that even topk variances are pretty minor. Forcing the model to hold this constraint will be harmful to the performance. SSMC loss uses margin to stop restricting the features, further relaxing the constraint and boosting the performance.

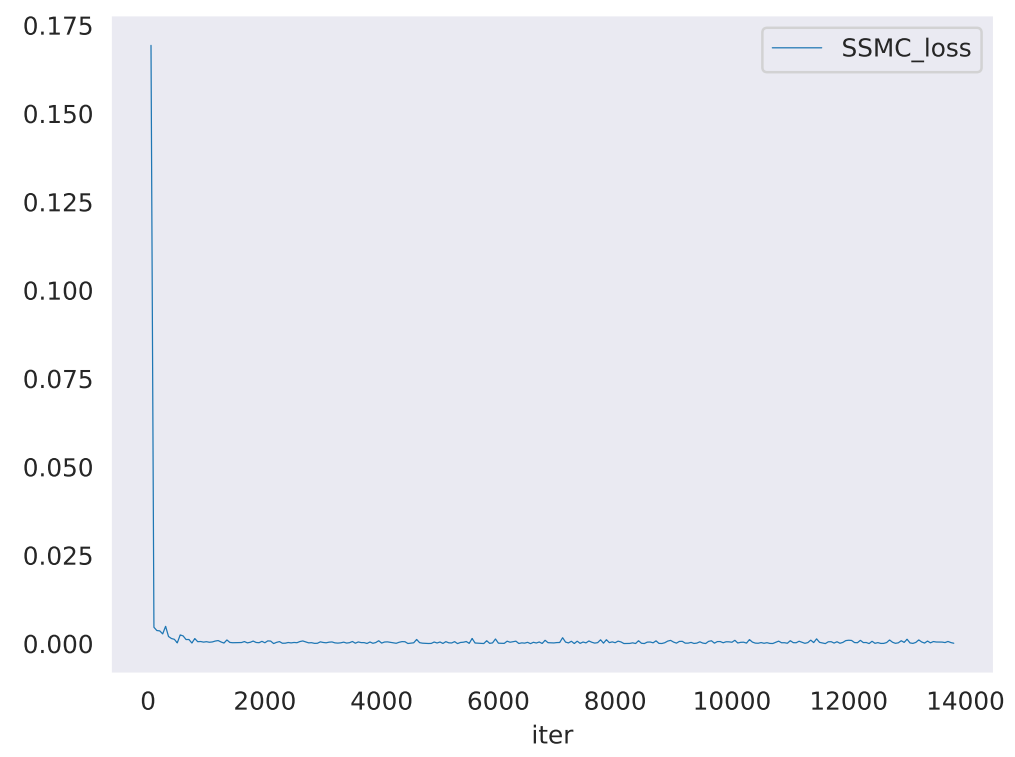

**Figure 8:** The loss curve of SSMC loss.

Furthermore, an experiment of $\lambda_{SSMC}$ is conducted as shown in Table 4. If $\lambda_{SSMC}$ increases to 10, the performance will gradually increase to 61.36%. If $\lambda_{SSMC}$ is too large, the performance will severely decline due to over-constraint.

Fig. 8 shows the curve of the SSMC loss. When the features from two streams become similar enough, the margin function will set them to 0. Thus, after a few iterations in the early stage, the SSMC loss will decrease very quickly and converge to a low level. The loss is stable and does not cause the collapse of learning.

### *4.2.4. Experiment of Mixup layer in ResNet-50*

As is shown in Table 5, the experiment illustrates the effect of different Mixup layers on the performance without SSMC loss. The experiment is implemented on the Faster R-CNN with ResNet-50 as the backbone. ResNet-50 contains 4 stages, and Mixup can be performed on each stage. It can be observed that when Mixup is utilized in the last three stages of ResNet, the highest performance of 60.32% can be obtained. However, increasing the number of Mixup layers can not guarantee performance improvement. From the fourth row to the sixth row, it can be concluded that Mixup on Medium depth leads to better performance compared with Mixup shallowly and Mixup deeply. Only Mixup on the stage-1 gets lower performance (58.45%) than only Mixup on other stages, which further proves that Mixup shallowly is not a reasonable choice. The reasons are listed as follows: (1) in the shallow layer, the domain manifold is less flattened, so it is difficult for linear interpolation to generate new domains; (2) in the deep layer, most of the irrelevant domain information is discarded according to Information Bottleneck [45], so the diversity of sampled domains is decreased.

### *4.2.5. Fewer source domains*

When the number of source domains decreases, does our method still perform well? Two more groups of experiments are set to find out the answer. From Table 6, our method outperforms DeepAll in all conditions, while the performance of DANN is lower than the performance of DeepAll in *type1-2* condition. However, the performance of DMCL does decline drastically when the number of source domains decreases because DMCL strongly relies on the input domain diversity to sample features on domain manifold by DMX.

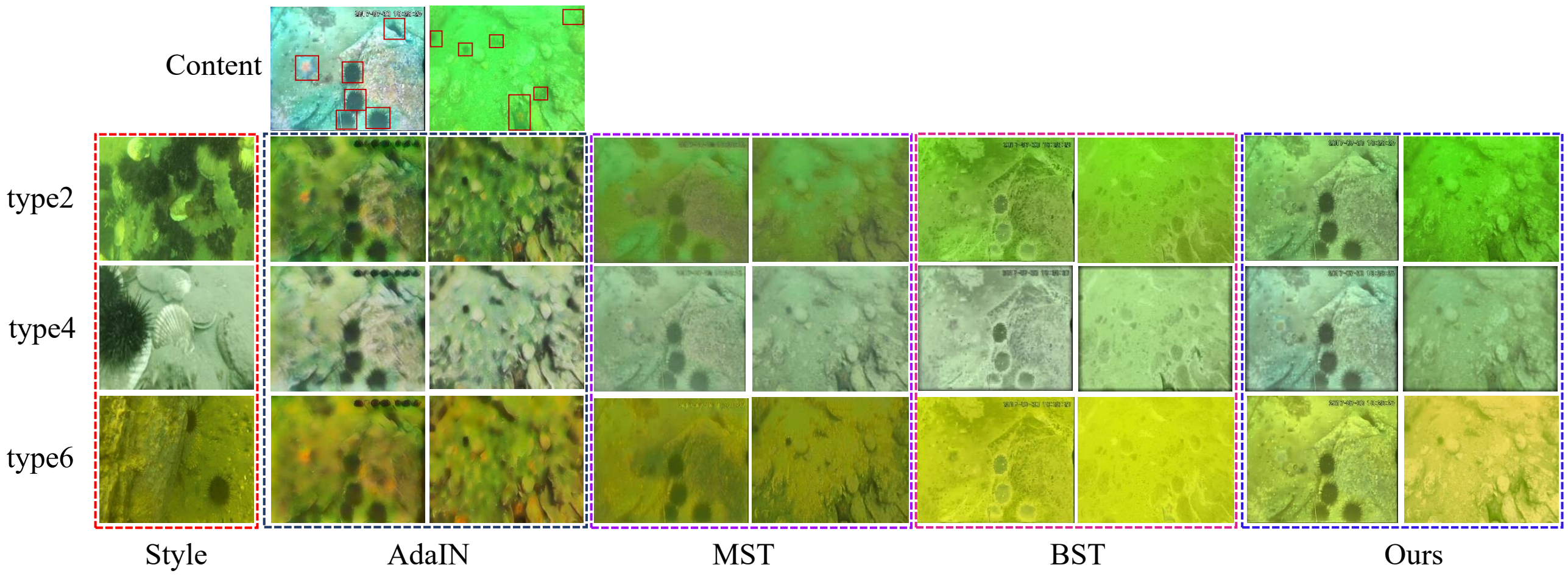

**Figure 9:** Qualitative comparison of our method against three state-of-the-art baselines on some challenging examples. Three very different domains of S-UODAC2020 are selected for comparison.

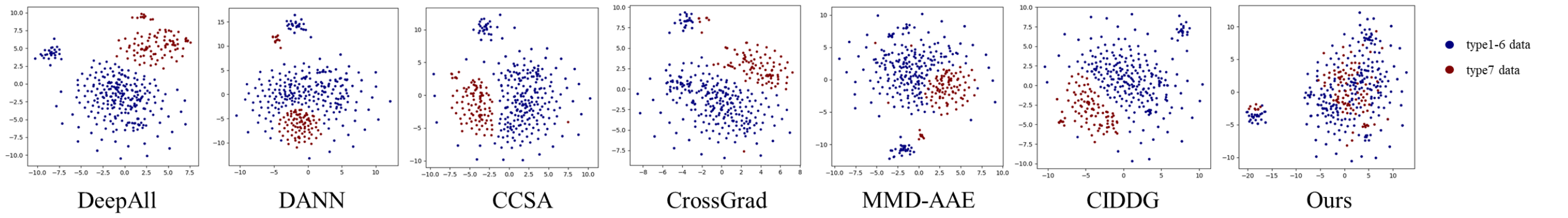

**Figure 10:** t-SNE visualization of different methods. Features are extracted from the last stage of ResNet with different methods. Blue points correspond to the source domain examples, while red points correspond to the target domain.

## 4.3. Experiments of Real-time Style Transfer Models

We make a comparison with three other real-time style transfer models: AdaIN [20], MST [21], and BST [56]. CBST is trained on an Nvidia GTX 1080Ti GPU with PyTorch implementation. The training dataset used as content images is UODAC2020 training set. One image from each source domain in S-UODAC2020 dataset is randomly selected as the style image, which means 6 fixed style images are used for training. The trained model is validated on UODAC2020 validation set. The input size is 512 × 512 both in training and inference. Adam is employed for optimization with learning rate set to 0.001, $\beta_1 = 0.9$ and $\beta_2 = 0.999$. CBST is trained for 10 epochs with batch size 8. For MST, the training details are the same as BST. Since AdaIN is open-sourced, its pre-trained model parameters can be directly used. BST has not been open-source yet, so we reimplement it on the same training setting as CBST, except that images in S-UODAC2020 dataset are randomly selected as the style images.

The qualitative comparison is shown in Fig. 9. The images AdaIN synthesizes are too stylized and not photo-realistic. MST blurs the objects, which may limit the performance of detection. BST even changes the contents of images, which may be because BST needs a large dataset and much more time to train. Using CIN instead of AdaIN helps fast convergence, and using Mask loss helps the images protect their important semantic information from essential changes. Thus, CBST generates clear images with semantic contents well-protected. Besides, from Table 7, when these style transfer models are applied to the training of Faster R-CNN, CBST obtains the highest performance, which is consistent with qualitative comparison, and it also achieves the fastest speed.

## 4.4. Visualization of t-SNE

T-SNE projection is leveraged to visualize the feature distributions. 300 images from source domains and 100 from the target domain are randomly selected and inputted into Faster R-CNN with different methods. The features of the last stage of ResNet-50 are projected on the 2D plane (Fig. 10). Blue points denote the data from source domains, while red points denote the data from the target domain. If the red points are well merged with the blue points, the model captures more domain-invariant features. From the subgraphs in Fig. 10, it can be quickly concluded that our

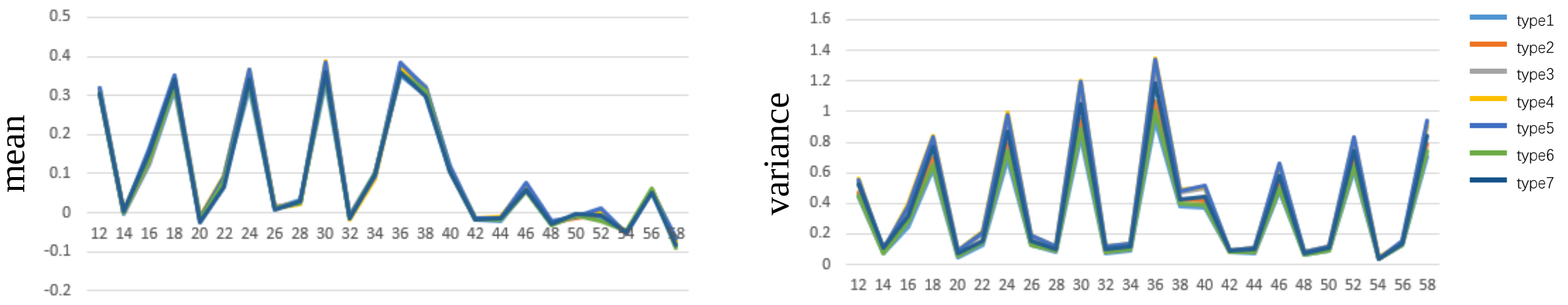

(a) DeepAll

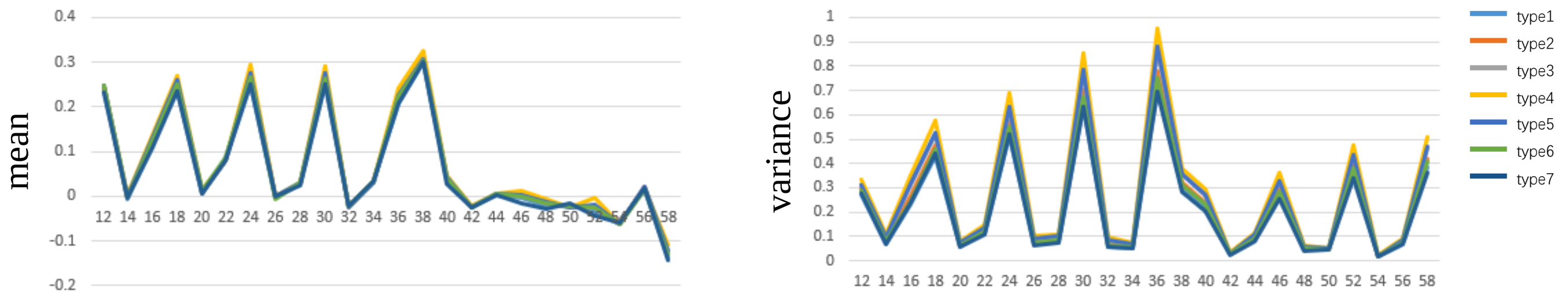

(b) Ours

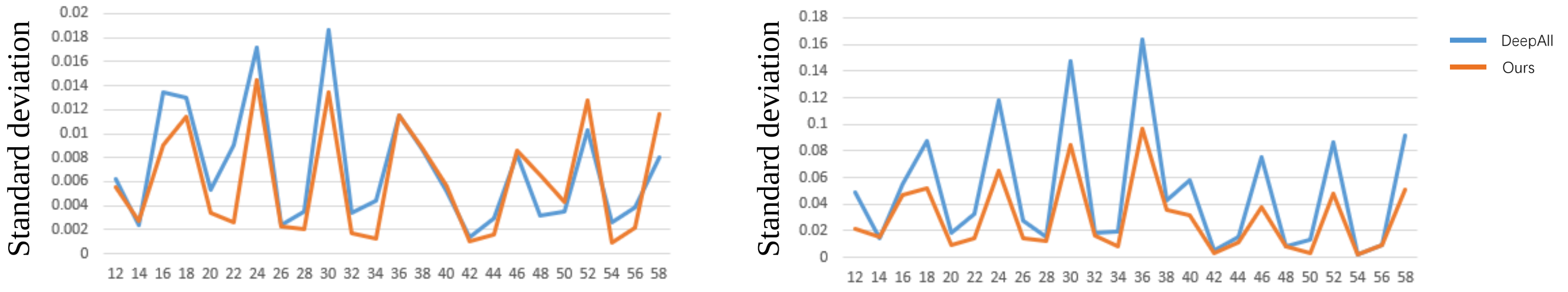

(c) Standard deviation of the features statistics

**Figure 11:** Visualization of the features statistics. X-axis denotes the layer indices. (a) Model trained on all source domains is analyzed. (b) Model trained with DMCL is analyzed. (c) The standard deviation of mean and variance of type1-7 for each layer.

method eliminates the effect of domain shift, while other methods somehow capture the spurious correlations.

## 4.5. Analysis of the features statistics

A lot of works about style transfer [21, 20] show that there are some connections between normalization parameters $(\gamma, \beta)$ and styles, which can be interpreted that the features statistics of models are related to domains. Inspired by the analysis in [52], the feature maps of each layer can be regarded as data that obey Gaussian distribution with specific mean and variance. In a domain-invariant model, the distribution of features from different domains will be close to each other, which means they will have similar means and variances in different domains. The statistics of the features are visualized in Fig. 11. 10 images are randomly selected from each domain, input into YOLOv3, and features are extracted from layer 12 to layer 58. In Fig. 11 (a-b), X-axis denotes layer indices, the left part denotes the mean of the features, and the right part denotes the variance of the features.

For Fig. 11 (a), the model trained on source domains is analyzed. Each line in Fig. 11 (a) represents the mean (left) or the variance (right) of the features from the 10 images in a certain layer. For Fig. 11 (b), the model trained with our method is analyzed. For Fig. 11 (c), we calculate the standard deviation of the mean (left) and variance (right) of *type1-7* for each layer. Supposed the model extracts domain-invariant features from different domains, the features extracted should have the same statistics. If the standard deviations of all domains are high, it is suggested that the model suffers from domain shift, while a slight standard deviation denotes the strong robustness to domain shift. From Fig. 11 (c), the standard deviation of features statistics of our methods (the orange line) is smaller than DeepAll (the blue line) in most of the layers. This visualization shows that our method helps the model learn more domain-invariant features.

## 4.6. Qualitative Comparisons

Fig. 12 shows the qualitative comparison between DMCL and other state-of-the-art methods on S-UODAC2020 dataset.

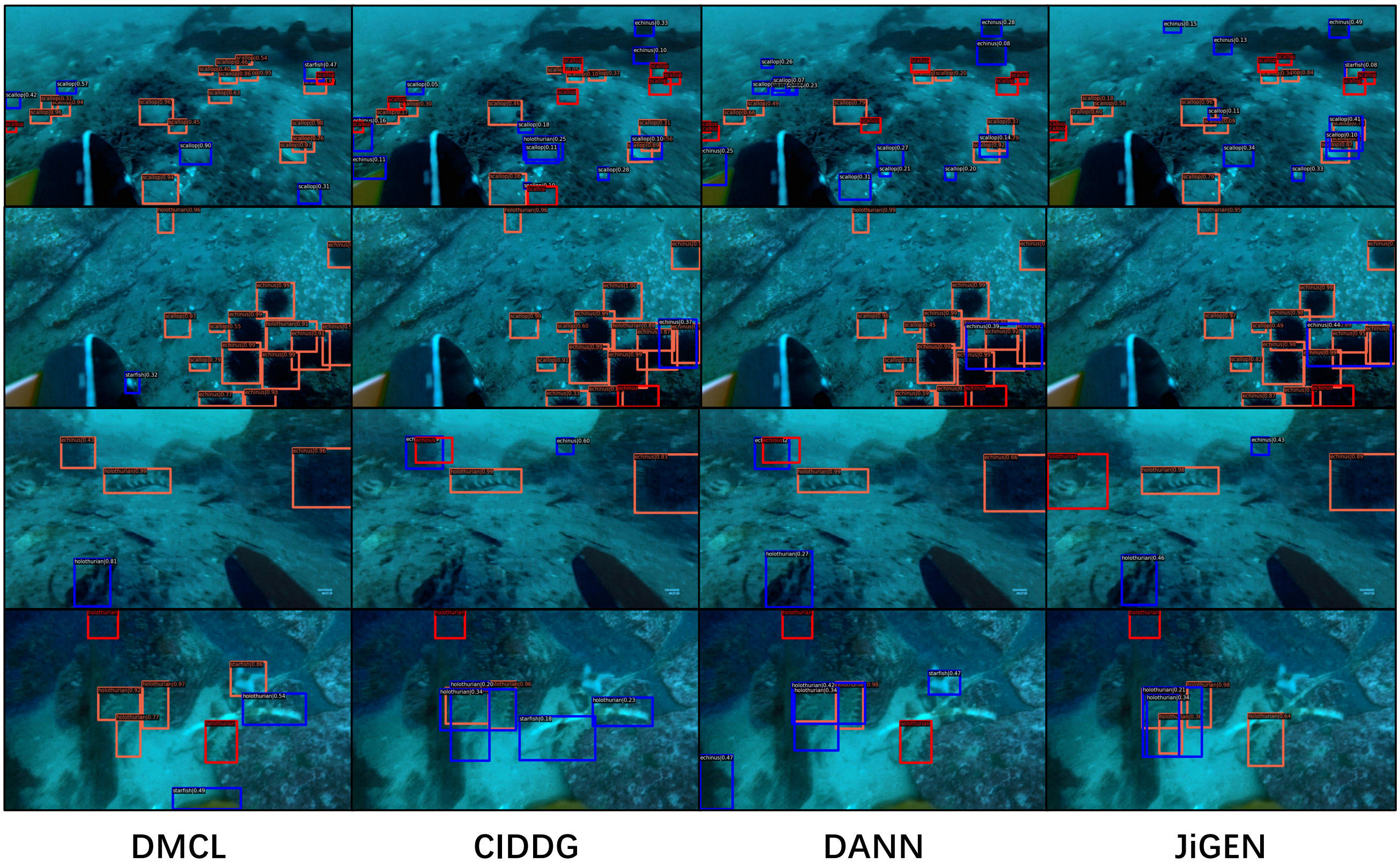

**Figure 12:** Qualitative comparison results on the S-UODAC2020 dataset. The orange boxes denote the matched predictions. The blue boxes denote the unmatched predictions. The red boxes denote the undetected ground truths.

We apply the detectors to some challenging cases, and the prediction score threshold is set to 0.05. For clarity, in each image, we visualize the prediction boxes with the top-k scores, and k is the number of the ground-truth boxes in the images. The orange boxes in the images denote a prediction box whose IoU with a certain ground truth is over 0.5 and higher than other predictions. The blue boxes denote the unmatched predictions whose IoUs with all ground truths are lower than 0.5. Besides, we also print the missed ground-truth boxes in red whose IoUs with all the predictions are lower than 0.5. Thus, more blue boxes in the images suggest lower precision, and more red boxes in the images suggest lower recall. As is shown in Fig. 12, compared with CIDDG, DANN and JiGEN, the proposed DMCL detects more objects (more orange boxes) with higher precision (fewer blue boxes). And DMCL misses fewer ground truths in cross-domain conditions (fewer red boxes). However, as shown in Fig. 13, DMCL is weak in detecting small objects. Small objects only occupy a small area of the feature maps. DMX will easily mix the feature of small objects with the background, making the feature not salient. Thus, it will lead to poor performance on small objects.

## 4.7. Comparison on other domain generalization datasets

### *4.7.1. Dataset*

Since there is no other accessible domain generalization dataset for detection, datasets of domain generalization for classification are chosen to further demonstrate the effectiveness of our proposed method. Our method is evaluated on two public domain generalization benchmark datasets: PACS and VLCS. PACS is a recent domain generalization benchmark for object recognition with larger domain discrepancy. It consists of seven object categories from four domains (Photo, Art Paintings, Cartoon, and Sketches datasets). The domain discrepancy among different datasets is larger than VLCS, making it more challenging. VLCS is a classic domain generalization benchmark for image classification, which includes five object categories from four domains (VOC PASCAL 2007, LabelMe, Caltech, and Sun datasets).

### *4.7.2. Comparison on PACS*

We follow the standard train/val/test splitting to train and evaluate our method. Besides, for PACS containing images that are not photorealistic (Art Paintings, Cartoon, and Sketches), our proposed photorealistic style transfer model CBST is no more effective. We decide to replace CBST with AdaIN [20]. Our method is evaluated with backbones

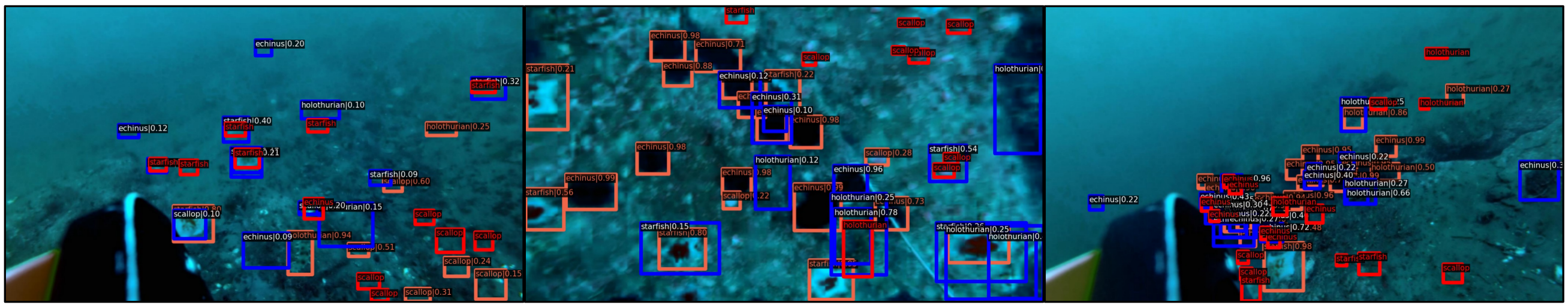

**Figure 13:** Some failure cases of DMCL.

**Table 8**
Domain Generalization Results on PACS Using Backbone ResNet-18.

| Method | Photo | Art painting | Cartoon | Sketch | Average |
|---|---|---|---|---|---|
| MASF [10] | 94.99 | 80.29 | 77.17 | 71.69 | 81.03 |
| Epi-FCR [24] | 93.90 | 82.10 | 77.00 | 73.00 | 81.50 |
| D-SAM [12] | 95.30 | 77.33 | 72.43 | 77.83 | 80.72 |
| JiGEN [3] | 96.03 | 79.42 | 75.25 | 71.35 | 80.51 |
| MetaReg [1] | 95.50 | **83.70** | 77.20 | 70.30 | 80.51 |
| EISNet [50] | 95.93 | 81.89 | 76.44 | 74.33 | 82.15 |
| L2A-OT [67] | **96.20** | 83.30 | 78.20 | 73.60 | 82.80 |
| **DMCL(Ours)** | 95.08 | 81.93 | **78.37** | **85.21** | **85.14** |

**Table 9**
Domain Generalization Results on VLCS Using Backbone AlexNet.

| Method | PASCAL | Labelme | Caltech | Sun | Average |
|---|---|---|---|---|---|
| CIDDG [32] | 64.38 | 63.06 | 88.83 | 62.10 | 69.59 |
| CCSA [40] | 67.10 | 62.10 | 77.00 | 59.10 | 70.15 |
| MMD-AAE [25] | 67.70 | 62.60 | 94.40 | 64.40 | 72.28 |
| D-SAM [12] | 58.59 | 56.95 | 91.75 | 60.84 | 67.03 |
| JiGEN [3] | **70.62** | 60.90 | **96.93** | 64.30 | **73.19** |
| **DMCL(Ours)** | 67.52 | **63.49** | 95.52 | **65.78** | 73.08 |

ResNet-18 pre-trained on ImageNet. Domain Mixup is performed on the features of the last three stages. The model is trained with a batch size of 64 for 100 epochs and optimized using Adam with a learning rate of 0.0005, $\beta_1$=0.9 and $\beta_2$=0.999. Cosine learning rate decay is leveraged during training. SSMC loss is applied in the last stage of ResNet-18.

Table 8 summaries the experimental results. Compared with other mainstream DG works, our method achieves SOTA performance. This result further shows the effectiveness of our method.

#### *4.7.3. Comparison on VLCS*

We follow the standard train/test splitting to train and evaluate our method. The validation set is split from the training set, with a proportion of 10%. Like PACS, our proposed photorealistic style transfer model CBST is replaced with AdaIN [20]. Our method is evaluated with backbones AlexNet. Domain Mixup is performed on the third ReLU operation and the last maxpooling layer. The model is trained with a batch size of 64 for 10 epochs, and optimize the model with Adam ($\beta_1$=0.9, $\beta_2$=0.999). Cosine learning rate decay is leveraged during training. We used a simple data augmentation protocol by randomly cropping the images to retain between 0.8-1, and randomly applying horizontal flipping. SSMC loss is applied on features just before the first fc-layer of AlexNet.

Table 9 summaries the experimental results. Unlike the excellent performance on the PACS dataset, the performance on the VLCS dataset does not reach the state of the art. Combined with the result in Table 6 and Table 8, the experiment further reveals the limitation of the proposed method. Our proposed DMCL strongly relies on the diversity of the source domain to sample new domains on the domain manifold. The diversity of source domains defines the area of the domain convex hull in the domain manifold. However, the domain diversity in the VLCS dataset is limited. Thus, the domains sampled in a small domain convex hull are limited. The performance improvement is less compared with other methods.

## 5. Conclusions

This paper focuses on domain generalization for underwater object detection, which is an interesting but understudied problem. DMCL is proposed based on two ideas: Domain Mixup and Contrastive Learning. Domain Mixup with style transfer model CBST can sample domain information on domain manifold by interpolation on the feature level, which enriches domain diversity of training data. SSMC loss can regularize the domain-specific information of the features. With our method, the detector will be domain-invariant. In experiments, our method achieves better performance on S-UODAC2020, PACS, and VLCS datasets compared with other domain generalization methods. Besides, further experiments prove the effectiveness and robustness of our method. We believe our method can promote the research of other domain adaptation and generalization tasks, such as autonomous driving in extreme weather (e.g., foggy or rainy days), and Re-ID in different light conditions (images of day and night).

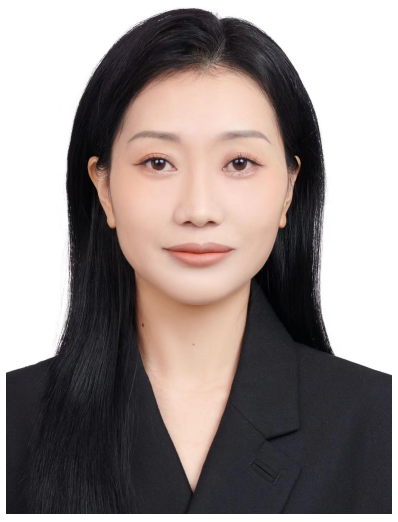

**Yang Chen** received the P.h.D. degree in mathematics in 2020 and majored in Differential Equations, Dynamical Systems and Optimal Control. Now she is doing research work in the computer application technology postdoctoral research station of the School of Information Engineering, Peking University, her research interests are computer vision and explainable deep learning.

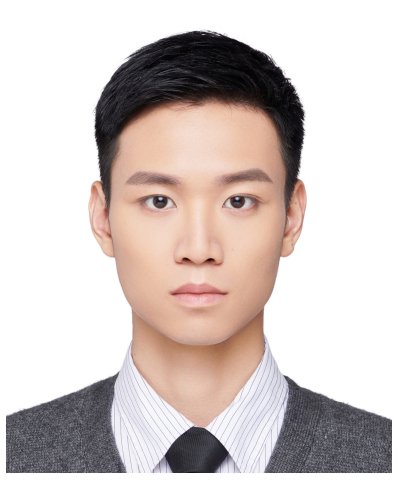

**Pinhao Song** received the B.E. degree in Mechanical Engineering in 2019, where he is currently pursuing a master's degree in computer applied technology in Peking University. His current research interests include underwater object detection, generic object detection, and domain generalization.

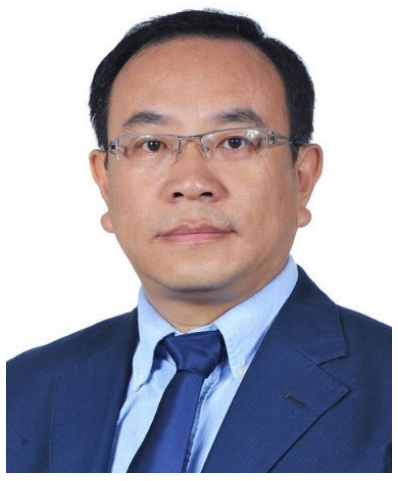

**Hong Liu** received the Ph.D. degree in mechanical electronics and automation in 1996. He serves as a Full Professor in the School of EE&CS, Peking University (PKU), China. Prof. Liu has been selected as Chinese Innovation Leading Talent supported by National High-level Talents Special Support Plan since 2013. Dr. Liu has published more than 200 papers and gained the Chinese National Aerospace Award, Wu Wenjun Award on Artificial Intelligence, Excellence Teaching Award, and Candidates of Top Ten Outstanding Professors in PKU. He has served as keynote speakers, co-chairs, session chairs, or PC members of many important international conferences, such as IEEE/RSJ IROS, IEEE ROBIO, IEEE SMC, and IIHMSP. Recently, Dr. Liu publishes many papers on international journals and conferences including TMM, TCSVT, TCYB, TALSP, TRO, PR, IJCAI, ICCV, CVPR, ICRA, IROS, etc.

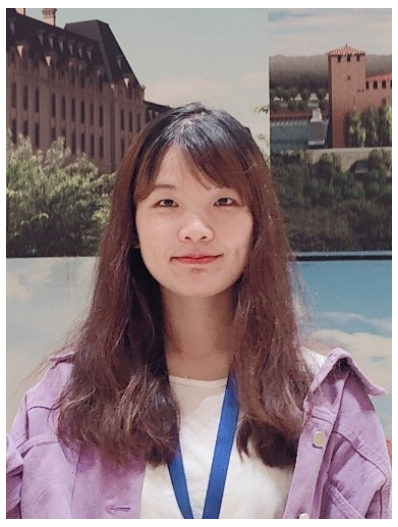

**Linhui Dai** received the B.E. degree in Information System and Information Management in 2018, and is currently working toward the Ph.D. degree with School of Electronics Engineering and Computer Science, Peking University, Beijing, China. Her current research interests include underwater object detection, open world object detection, and salient object detection.

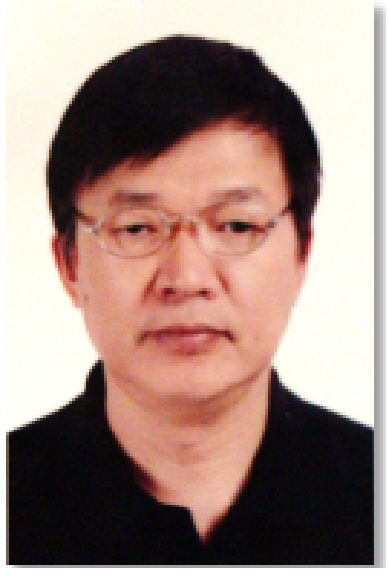

**Xiaochuan Zhang** is a professor in Chongqing University of Technology. The main research areas are machine games, intelligent robots, computational intelligence and software engineering. Currently, he is the vice president of school of Artificial Intelligence of Chongqing University of Technology, director of the Institute of Artificial Intelligence Systems, executive director of the Chinese Society for Artificial Intelligence and the director of the Machine Game Professional Committee and the executive director and deputy secretary-general of the Chongqing Artificial Intelligence Society, National first-class professional leader (software engineering), won 4 provincial and ministerial science and technology and teaching awards.

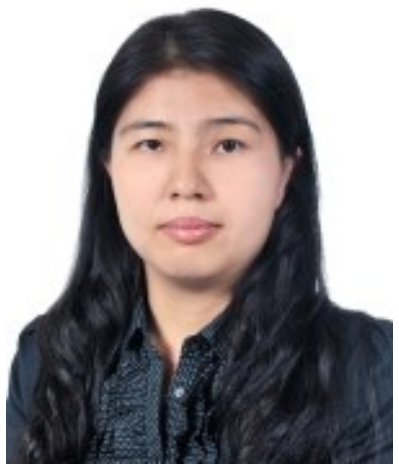

**Runwei Ding** received the Ph.D degree in computer applied technology from Peking University in 2019. She has published articles in the International Journal of Advanced Robotic Systems, the IEEE International Conference on Acoustics, Speech, and Signal Processing, and the IEEE International Conference on Image Processing. Her research interests include action recognition and localization.

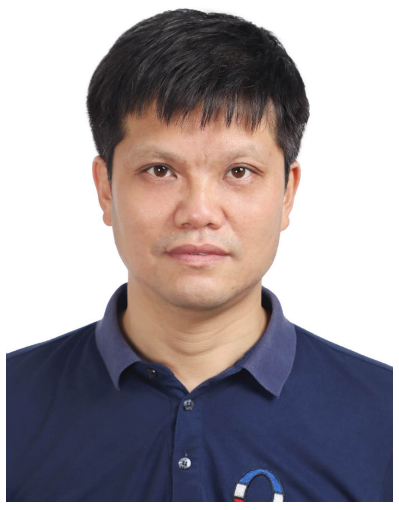

**Shengquan Li** received the B.S. degree from National University of Defense Technology, Changsha, China, in 1994, and M.E. degree from Dalian Naval Academy, Dalian, China, in 2006, and Ph.D. degree from Naval University of Engineer, Wuhan, China, in 2013. He was with college of Army Missile as an Assistant Professor from 1994 to 1996. He was with Hydrographic and Charting Institute as an Engineer from 1996 to 2007 and a Senior Engineer from 2007 to 2016. He was a Professor of college of Underwater Acoustic Engineering with Harbin Engineering University from 2016 to 2019. He is currently a Professor with the Peng Cheng Laboratory (PCL), Shenzhen. He is also the Director of the Institute of Maritime Information Technology at PCL. His research interests are measurement techniques and information expression of ocean spatial structure, especially in GNSS, sonar, and GIS.